%% file: emnlp2023.tex
\pdfoutput=1

\documentclass[11pt]{article}

\usepackage{EMNLP2023}

\usepackage{times}
\usepackage{latexsym}
\usepackage{arydshln}
\usepackage{graphicx}
\usepackage{subcaption}
\usepackage{hyperref}
\usepackage{booktabs,arydshln}
\usepackage{amsmath}
\usepackage{enumitem}
\usepackage{multirow}
\usepackage{cleveref}
\usepackage[export]{adjustbox}
\usepackage[utf8]{inputenc}
\usepackage{xurl}
\usepackage{tabularray}
\usepackage{siunitx}
\usepackage{amssymb}
\usepackage{pifont}
\newcommand{\cmark}{\ding{51}}%
\newcommand{\xmark}{\ding{55}}%

\newcommand{\circa}{{\raise.17ex\hbox{$\scriptstyle\sim$}}}

\newcommand{\ourmodel}[0]{PHD }
\newcommand{\ourmodels}[0]{PHD's }
\newcommand{\ourmodelnospace}[0]{PHD}

\usepackage{array}
\newcolumntype{L}[1]{>{\raggedright\let\newline\\\arraybackslash\hspace{0pt}}m{#1}}
\newcolumntype{C}[1]{>{\centering\let\newline\\\arraybackslash\hspace{0pt}}m{#1}}
\newcolumntype{R}[1]{>{\raggedleft\let\newline\\\arraybackslash\hspace{0pt}}m{#1}}

\newcommand\nnfootnote[1]{%
  \begin{NoHyper}
  \renewcommand\thefootnote{}\footnote{#1}%
  \addtocounter{footnote}{-1}%
  \end{NoHyper}
}

\NewDocumentCommand\githubicon{}{\includegraphics[scale=0.025]{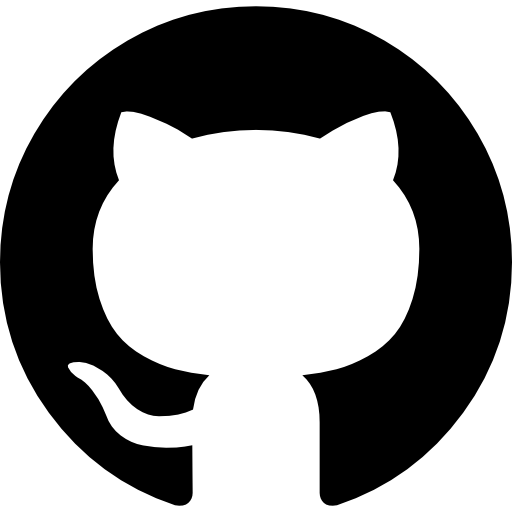}
}

\usepackage[T1]{fontenc}

\usepackage[utf8]{inputenc}

\usepackage{microtype}

\title{PHD: Pixel-Based Language Modeling of Historical Documents}

\author{Nadav Borenstein \ \ Phillip Rust \ \ Desmond Elliott \ \ Isabelle Augenstein\\
Department of Computer Science, University of Copenhagen\\
\texttt{\{nadav.borenstein, p.rust, de, augenstein\}@di.ku.dk} }

\begin{document}
\maketitle

\begin{abstract}
\input{Modules/0-Abstract}

\nnfootnote{\textcolor{Bittersweet!60}{*This paper shows dataset samples that are racist in nature}}
\end{abstract}

\everypar{\looseness=-1}

\section{Introduction}
\label{sec:introduction}
\input{Modules/1-Introduction}

\section{Background}
\label{sec:related_work}
\input{Modules/2-Related-work}

\section{Model}
\label{sec:model}

\input{Modules/3-model}

\section{Training a Pixel-Based Historical LM}
\label{sec:pretraining}

\input{Modules/4-Pretraining}

\section{Training for Downstream NLU Tasks}
\label{sec:finetuning}
\input{Modules/5-Finetuning}

\section{Conclusion}
\label{sec:conclusion}
\input{Modules/6-Conclusion}

\section*{Acknowledgements}
This research was partially funded by a DFF Sapere Aude research leader grant under grant agreement No 0171-00034B, the Danish-Israeli Study Foundation in Memory of Josef and Regine Nachemsohn, the Novo Nordisk Foundation (grant NNF 20SA0066568), as well as by a research grant (VIL53122) from VILLUM FONDEN. The research was also supported by the Pioneer Centre for AI, DNRF grant number P1.

\newpage

\section*{Limitations}
\label{sec:limitations}
\input{Modules/7-Limitations}

\bibliography{anthology,custom}
\bibliographystyle{acl_natbib}

\newpage

\appendix

\input{Modules/8-Appendix}

\end{document}

%% file: Modules/0-Abstract.tex
The digitisation of historical documents has provided historians with unprecedented research opportunities. Yet, the conventional approach to analysing historical documents involves converting them from images to text using OCR, a process that overlooks the potential benefits of treating them as images and introduces high levels of noise. 
To bridge this gap, we take advantage of recent advancements in pixel-based language models trained to reconstruct masked patches of pixels instead of predicting token distributions. 
Due to the scarcity of real historical scans, we propose a novel method for generating synthetic scans to resemble real historical documents. We then pre-train our model, \ourmodelnospace, on a combination of synthetic scans and real historical newspapers from the 1700-1900 period.
Through our experiments, we demonstrate that \ourmodel exhibits high proficiency in reconstructing masked image patches and provide evidence of our model's noteworthy language understanding capabilities. Notably, we successfully apply our model to a historical QA task, highlighting its utility in this domain.

%% file: Modules/1-Introduction.tex
\begin{figure*}[t]
    \centering
    \begin{subfigure}{0.25\textwidth}
        \centering
        \fboxsep=0pt
        \fbox{
        \includegraphics[width=\textwidth, center]{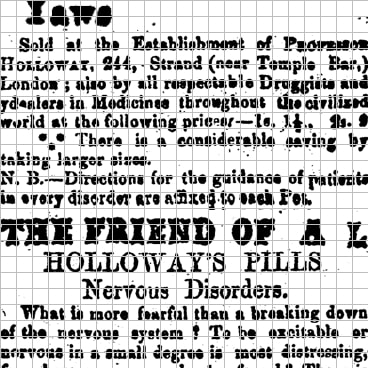}
        }
        \caption{Input example.}
        \label{fig:input_example}
      \end{subfigure} \hspace{0.7 cm}
    \begin{subfigure}{0.25\textwidth}
        \centering
        \fboxsep=0pt
        \fbox{
        \includegraphics[width=\textwidth, center]{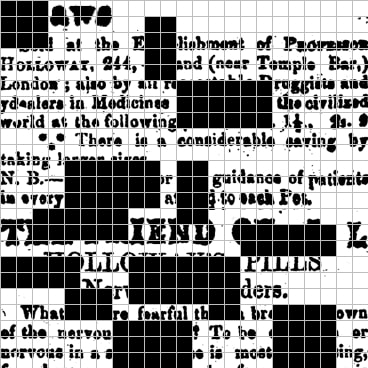}
        }
        \caption{Masking the input.}
        \label{fig:input_masking}
      \end{subfigure} \hspace{0.7 cm}
        \begin{subfigure}{0.25\textwidth}
        \centering
        \fboxsep=0pt
        \fbox{
        \includegraphics[width=\textwidth, center]{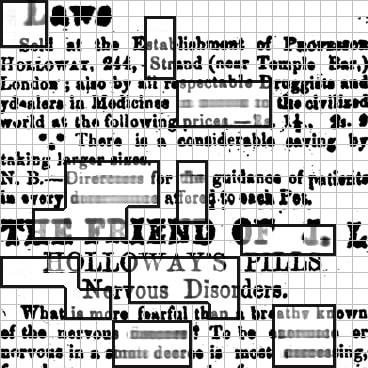}
        }
        \caption{Model predictions.}
        \label{fig:model_predictions}
      \end{subfigure}

    \caption{Our proposed model, \ourmodelnospace. The model is trained to reconstruct the original image (a) from the masked image (b), resulting in (c). The grid represents the 16 $\times$ 16 pixels patches that the inputs are broken into.}
    \label{fig:proposed_model}%
\end{figure*}

Recent years have seen a boom in efforts to digitise historical documents in numerous languages and sources \cite{alma99122601258605763, delpher, moss2009guides}, leading to a transformation in the way historians work. Researchers are now able to expedite the analysis process of vast historical corpora using NLP tools, thereby enabling them to focus on interpretation instead of the arduous task of evidence collection \cite{laite2020emmet, gerritsen2012scales}.

The primary step in most NLP tools tailored for historical analysis involves Optical Character Recognition (OCR). However, this approach poses several challenges and drawbacks. First, OCR strips away any valuable contextual meaning embedded within non-textual elements, such as page layout, fonts, and figures.\footnote{Consider, for example, the visual data that is lost by processing the newspaper page in Fig \ref{fig:full_newspaper_page} in App \ref{sec:additional_results} as text.}
Moreover, historical documents present numerous challenges to OCR systems. This can range from deteriorated pages, archaic fonts and language, the presence of non-textual elements, and occasional deficiencies in scan quality (e.g., blurriness), all of which contribute to the introduction of additional noise. Consequently, the extracted text is often riddled with errors at the character level \cite{robertson-goldwater-2018-evaluating, bollmann-2019-large}, which most large language models (LLMs) are not tuned to process. Token-based LLMs are especially sensitive to this, as the discrete structure of their input space cannot handle well the abundance of out-of-vocabulary words that characterise OCRed historical documents \cite{rust2022language}. Therefore, while LLMs have proven remarkably successful in modern domains, their performance is considerably weaker when applied to historical texts \citep[][\textit{inter alia}]{jdmdh:9690, baptiste2021transferring}.
Finally, for many languages, OCR systems either do not exist or perform particularly poorly. As training new OCR models is laborious and expensive \cite{li2021trocr}, the application of NLP tools to historical documents in these languages is limited.

This work addresses these limitations by taking advantage of recent advancements in pixel-based language modelling, with the goal of constructing a general-purpose, image-based and OCR-free language encoder of historical documents. Specifically, we adapt PIXEL \cite{rust2022language}, a language model that renders text as images and is trained to reconstruct masked patches instead of predicting a distribution over tokens. PIXEL's training methodology is highly suitable for the historical domain, as (unlike other pixel-based language models) it does not rely on a pretraining dataset composed of instances where the image and text are aligned. Fig \ref{fig:proposed_model} visualises our proposed training approach.

Given the paucity of large, high-quality datasets comprising historical scans, we pretrain our model using a combination of 1) synthetic scans designed to resemble historical documents faithfully, produced using a novel method we propose for synthetic scan generation; and 2) real historical English newspapers published in the Caribbeans in the 18th and 19th centuries. The resulting pixel-based language encoder, \ourmodel (\textbf{P}ixel-based model for \textbf{H}istorical \textbf{D}ocuments), is subsequently evaluated based on its comprehension of natural language and its effectiveness in performing Question Answering from historical documents. 

We discover that \ourmodel displays impressive reconstruction capabilities, being able to correctly predict both the form and content of masked patches of historical newspapers (\S \ref{sec:pretraining_results}). We also note the challenges concerning quantitatively evaluating these predictions. We provide evidence of our model's noteworthy language understanding capabilities while exhibiting an impressive resilience to noise. Finally, we demonstrate the usefulness of the model when applied to the historical QA task (\S\ref{sec:finetuning_results}).

To facilitate future research, we provide the dataset, models, and code at \githubicon \url{https://github.com/nadavborenstein/pixel-bw}.

%% file: Modules/2-Related-work.tex
\subsection{NLP for Historical Texts}


Considerable efforts have been invested in improving both OCR accuracy \cite{li2021trocr, tesseract} and text normalisation techniques for historical documents \cite{spell_correction_2017, robertson-goldwater-2018-evaluating, text_normalization_2018b, bollmann-2019-large, spell_correction_2021}. This has been done with the aim of aligning historical texts with their modern counterparts.
However, 
these methods are not without flaws \cite{robertson-goldwater-2018-evaluating, bollmann-2019-large}, and any errors introduced during these preprocessing stages can propagate to downstream tasks \cite{robertson-goldwater-2018-evaluating, hill2019quantifying}. As a result, historical texts remain a persistently challenging domain for NLP research \cite{historical_event_extraction_2021, de-toni-etal-2022-entities, nadav2023karolina}. Here, we propose a novel approach to overcome the challenges associated with OCR in historical material, by employing an image-based language model capable of directly processing historical document scans and effectively bypassing the OCR stage.

\subsection{Pixel-based Models for NLU}
\label{sec2:pixel_based_models}
Extensive research has been conducted on models for processing text embedded in images. Most existing approaches incorporate OCR systems as an integral part of their inference pipeline \cite{appalaraju2021docformer, li2021selfdoc, delteil2022matrix}. These approaches employ multimodal architectures where the input consists of both the image and the output generated by an OCR system.

Recent years have also witnessed the emergence of OCR-free approaches for pixel-based language understanding. \citet{kim2022ocr} introduce Donut, an image-encoder-text-decoder model for document comprehension. Donut is pretrained with the objective of extracting text from scans, a task they refer to as ``pseudo-OCR''. Subsequently, it is finetuned on various text generation tasks, reminiscent of T5 \cite{roberts-etal-2020-much}. While architecturally similar to Donut, Dessurt \cite{davis2022end} and Pix2Struct \cite{lee2022pix2struct} were pretrained by masking image regions and predicting the text in both masked and unmasked image regions. Unlike our method, all above-mentioned models predict in the text space rather than the pixel space. This presupposes access to a pretraining dataset comprised of instances where the image and text are aligned. However, this assumption cannot hold for historical NLP since OCR-independent ground truth text for historical scans is, in many times, unprocurable and cannot be used for training purposes. 

Text-free models that operate at the pixel level for language understanding are relatively uncommon. One notable exception is \citet{li2022dit}, which utilises  Masked Image Modeling for pretraining on document patches. Nevertheless, their focus lies primarily on tasks that do not necessitate robust language understanding, such as table detection, document classification, and layout analysis. PIXEL \cite{rust2022language}, conversely, is a text-free pixel-based language model that exhibits strong language understanding capabilities, making it the ideal choice for our research. The subsequent section will delve into a more detailed discussion of PIXEL and how we adapt it to our task.

%% file: Modules/3-Model.tex
\textbf{PIXEL} We base \ourmodel on PIXEL, a pretrained pixel-based encoder of language. PIXEL has three main components: A text renderer that draws texts as images, a pixel-based encoder, and a pixel-based decoder. The training of PIXEL is analogous to BERT \cite{devlin-etal-2019-bert}. During pretraining, input strings are rendered as images, and the encoder and the decoder are trained jointly to reconstruct randomly masked image regions from the unmasked context. During finetuning, the decoder is replaced with a suitable classification head, and no masking is performed. The encoder and decoder are based on the ViT-MAE architecture \cite{he2022masked} and work at the patch level. That is, the encoder breaks the input image into patches of \num{16} $\times$ \num{16} pixels and outputs an embedding for each patch. The decoder then decodes these patch embeddings back into pixels. Therefore, random masking is performed at the patch level as well.


\paragraph{\ourmodelnospace} We follow the same approach as PIXEL's pretraining and finetuning schemes. However, PIXEL's intended use is to process texts, not natural images. That is, the expected input to PIXEL is a string, not an image file. In contrast, we aim to use the model to encode real document scans. Therefore, we make several adaptations to PIXEL's training and data processing procedures to make it compatible with our use case (\S\ref{sec:pretraining} and \S\ref{sec:finetuning}). 

Most crucially, we alter the dimensions of the model's input: The text renderer of PIXEL renders strings as a long and narrow image with a resolution of \num{16} $\times$ \num{8464} pixels (corresponding to \num{1} $\times$ \num{529} patches), such that the resulting image resembles a ribbon with text. Each input character is set to be not taller than \num{16} pixels and occupies roughly one patch. However, real document scans cannot be represented this way, as they have a natural two-dimensional structure and irregular fonts, 
as Fig \ref{fig:input_example} demonstrates (and compare to Fig \ref{fig:pixel_input} in App \ref{sec:additional_results}). Therefore, we set the input size of \ourmodel to be \num{368} $\times$ 
\num{368} pixels (or \num{23} $\times$ \num{23} patches). 


%% file: Modules/4-Pretraining.tex
\begin{table}[t]
    \centering
    \resizebox{0.48\textwidth}{!}{%
    \fontsize{10}{10}\selectfont
    \sisetup{table-format = 3.2, group-minimum-digits=3}
    \begin{tabular}{p{1.9cm} r R{1.4cm} R{1.3cm}}
        \toprule
         \textbf{Source} & \textbf{$\#$Issues} & \textbf{$\#$Train Scans} & \textbf{$\#$Test Scans} \\ 
        \midrule 
        
        Caribbean & \multirow{2}{*}{\num{7487}} & \multirow{2}{*}{\num{1675172}} & \multirow{2}{*}{\num{87721}}\\
        Project & & &\\\addlinespace[0.3em]
        Danish Royal & \multirow{2}{*}{\num{5661}} & \multirow{2}{*}{\num{300780}} & \multirow{2}{*}{\num{15159}}\\
        Library & & & \\ \midrule 
        
        Total & \num{13148} & \num{1975952} & \num{102880}\\
        \bottomrule
    \end{tabular}
    }
    
    \caption{Statistics of the newspapers  dataset.}
    \label{tab:dataset_statistics}
\end{table}

We design \ourmodel to serve as a general-purpose, pixel-based language encoder of historical documents. Ideally, \ourmodel should be pretrained on a large dataset of scanned documents from various historical periods and different locations. 
However, large, high-quality datasets of historical scans are not easily obtainable. 
Therefore, we propose a novel method for generating historical-looking artificial data from modern corpora (see \autoref{sec:fake_scans}). We adapt our model to the historical domain by continuously pretraining it on a medium-sized corpus of real historical documents. Below, we describe the datasets and the pretraining process of the model.

\subsection{Artificially Generated Pretraining Data}
\label{sec:fake_scans}

\begin{figure*}[t]
    \centering
    \begin{subfigure}{0.25\textwidth}
        \centering
        \fboxsep=0pt
        \fbox{
        \includegraphics[width=\textwidth, center]{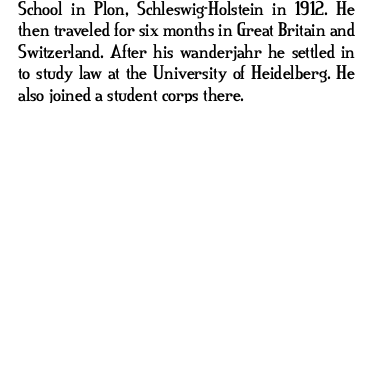}
        }
        \caption{Embedding one paragraph.}
        \label{fig:pretraining_sample_first_paragraph}
      \end{subfigure} 
      \hspace{0.7 cm}
    \begin{subfigure}{0.25\textwidth}
        \centering
        \fboxsep=0pt
        \fbox{
        \includegraphics[width=\textwidth, center]{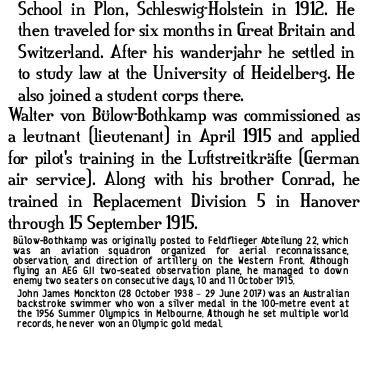}
        }
        \caption{Adding more paragraphs.}
        \label{fig:pretraining_sample_all_paragraphs}
      \end{subfigure} 
      \hspace{0.7 cm}
        \begin{subfigure}{0.25\textwidth}
        \centering
        \fboxsep=0pt
        \fbox{
        \includegraphics[width=\textwidth, center]{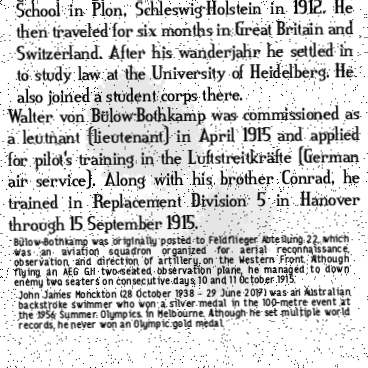}
        }
        \caption{Adding noise.}
        \label{fig:pretraining_sample_with_augmentations}
      \end{subfigure}

    \caption{Process of generating a single artificial scan. Refer to \S\ref{sec:fake_scans} for detailed explanations.}
    \label{fig:pretraining_sample}%
\end{figure*}

Our pretraining dataset consists of artificially generated scans of texts from the same sources that BERT used, namely the BookCorpus \cite{Zhu_2015_ICCV} and the English Wikipedia.\footnote{We use the version ``20220301.en'' hosted on \href{https://huggingface.co/datasets/wikipedia}{\url{huggingface.co/datasets/wikipedia}.}} We generate the scans as follows.

We generate dataset samples on-the-fly, adopting a similar approach as \citet{davis2022end}. First, we split the text corpora into paragraphs, using the new-line character as a delimiter. From a paragraph chosen at random, we pick a random spot and keep the text spanning from that spot to the paragraph's end. We also sample a random font and font size from a pre-defined list of fonts (from \citet{davis2022end}). The text span and the font are then embedded within an HTML template using the Python package Jinja,\footnote{\href{https://jinja.palletsprojects.com/en/3.1.x/}{\url{jinja.palletsprojects.com/en/3.1.x}}} set to generate a Web page with dimensions that match the input dimension of the model. Finally, we use the Python package WeasyPrint\footnote{\href{https://weasyprint.org}{\url{weasyprint.org}}} to render the HTML file as a PNG image. Fig \ref{fig:pretraining_sample_first_paragraph} visualises this process' outcome.

In some cases, if the text span is short or the selected font is small, the resulting image contains a large empty space (as in Fig \ref{fig:pretraining_sample_first_paragraph}). When the empty space within an image exceeds 10\%, a new image is generated to replace the vacant area. We create the new image by randomly choosing one of two options. In 80\% of the cases, we retain the font of the original image and select the next paragraph. In 20\% of the cases, a new paragraph and font are sampled. This pertains to the common case where a historical scan depicts a transition of context or font (e.g., Fig \ref{fig:input_example}). This process can repeat multiple times, resulting in images akin to Fig  \ref{fig:pretraining_sample_all_paragraphs}. 

Finally, to simulate the effects of scanning ageing historical documents, we degrade the image by adding various types of noise, such as blurring, rotations, salt-and-pepper noise and bleed-through effect (see Fig \ref{fig:pretraining_sample_with_augmentations} and Fig \ref{fig:artificial_samples_extra} in App \ref{sec:additional_results} for examples). App \ref{app:dataset_generation} enumerates the full list of the degradations and augmentations we use.

\subsection{Real Historical Scans}
\label{sec:real_scans}

\begin{figure*}[t]
    \centering
        \includegraphics[trim={0.3cm 0.3cm 0.3cm 0.3cm},clip, width=0.855\textwidth]{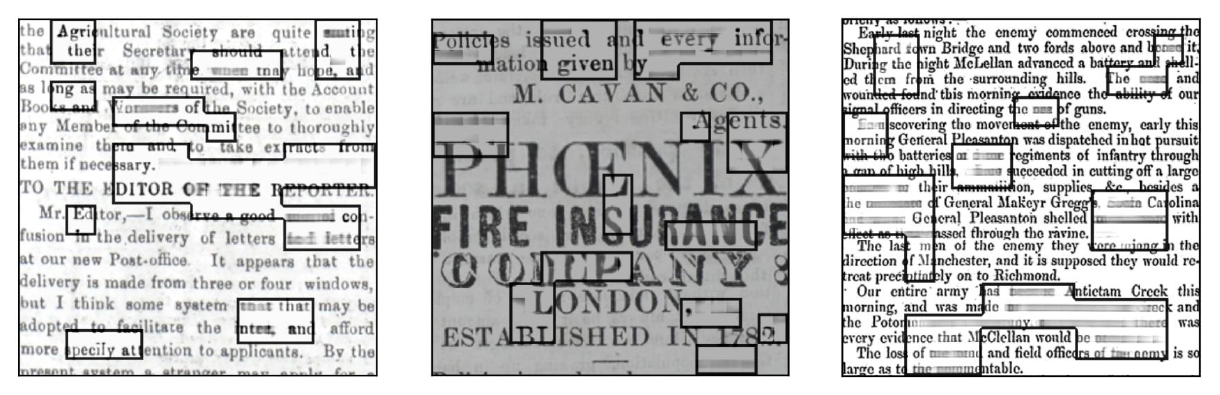}
         \caption{Examples of some image completions made by \ourmodel. Masked regions marked by dark outlines.}
         \label{fig:simple_completions}
\end{figure*}

We adapt \ourmodel to the historical domain by continuously pretraining it on a medium-sized corpus of scans of real historical newspapers. Specifically, we collect newspapers written in English from the  ``Caribbean Newspapers, 1718--1876'' database,\footnote{\href{https://www.readex.com/products/caribbean-newspapers-series-1-1718-1876-american-antiquarian-society}{\url{readex.com/products/caribbean-newspapers-series-1-1718-1876-american-antiquarian-society}}} the largest collection of Caribbean newspapers from the 18th--19th century available online. We extend this dataset with English-Danish newspapers published between 1770--1850 in the Danish Caribbean colony of Santa Cruz (now Saint Croix) downloaded from the Danish Royal Library's website.\footnote{\href{https://www2.statsbiblioteket.dk/mediestream/}{\url{statsbiblioteket.dk/mediestream}}} See Tab~\ref{tab:dataset_statistics} for details of dataset sizes. While confined in its geographical and temporal context, this dataset offers a rich diversity in terms of content and format, rendering it an effective test bed for evaluating \ourmodelnospace.

Newspaper pages are converted into a \num{368} $\times$ \num{368} pixels crops using a sliding window approach over the page's columns. This process is described in more detail in App \ref{app:dataset_generation}. We reserve 5\% of newspaper issues for validation, using the rest for training. See Fig \ref{fig:real_dataset_samples} in App \ref{sec:additional_results} for dataset examples.

\subsection{Pretraining Procedure}
\label{sec:pretraining_procedure}

Like PIXEL, the pretraining objective of \ourmodel is to reconstruct the pixels in masked image patches. We randomly occlude 28\% of the input patches with 2D rectangular masks. We uniformly sample their width and height from $[2, 6]$ and $[2, 4]$ patches, respectively, and then place them in random image locations (See Fig \ref{fig:input_masking} for an example). Training hyperparameters can be found in App \ref{app:training}.

\subsection{Pretraining Results}
\label{sec:pretraining_results}

\begin{figure*}[t]
    \centering
    \begin{subfigure}{0.25\textwidth}
        \fboxsep=0pt
    \fbox{
        \includegraphics[width=\textwidth]{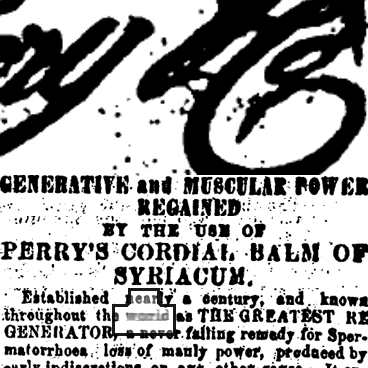}
        }
        \caption{world}
    \end{subfigure}
    \hspace{0.7cm}
    \begin{subfigure}{0.25\textwidth}
        \fboxsep=0pt
    \fbox{
        \includegraphics[width=\textwidth]{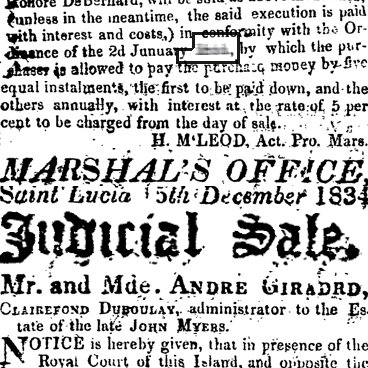}
        }
        \caption{1893}
    \end{subfigure}
    \hspace{0.7cm}
    \begin{subfigure}{0.25\textwidth}
        \fboxsep=0pt
    \fbox{
        \includegraphics[width=\textwidth]{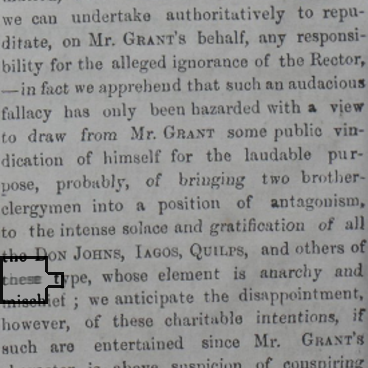}
        }
        \caption{every}
        \label{fig:every}
    \end{subfigure}

    \caption{Single word completions made by our model. Figure captions depict the missing word. Fig (a) depicts a successful reconstruction, whereas Fig (b) and (c) represent fail-cases.}
    \label{fig:single_word_completion}
\end{figure*}

\textbf{Qualitative Evaluation.} We begin by conducting a qualitative examination of the predictions made by our model. Fig \ref{fig:simple_completions} presents a visual representation of the model's predictions on three randomly selected scans from the test set of the Caribbean newspapers dataset  (for additional results on other datasets, refer to Fig \ref{fig:completions_extra} App \ref{sec:additional_results}). From a visual inspection,
it becomes evident that the model accurately reconstructs the fonts and structure of the masked regions. However, the situation is less clear when it comes to predicting textual content. Similar to \citet{rust2022language}, unsurprisingly, prediction quality is high and the results are sharp for smaller masks and when words are only partially obscured. However, as the completions become longer, the text quality deteriorates, resulting in blurry text.  It is important to note that evaluating these blurry completions presents a significant challenge. Unlike token-based models, where the presence of multiple words with high, similar likelihood can easily be detected by examining the discrete distribution, this becomes impossible with pixel-based models. In pixel-based completions, high-likelihood words may overlay and produce a blurry completion. Clear completions are only observed when a single word has a significantly higher probability compared to others. This limitation is an area that we leave for future work. 

We now move to analyse \ourmodels ability to fill in single masked words. We randomly sample test scans and OCRed them using Tesseract.\footnote{\href{http://github.com/tesseract-ocr/tesseract}{\url{github.com/tesseract-ocr/tesseract}}} Next, we randomly select a single word from the OCRed text and use Tesseract's word-to-image location functionality to (heuristically) mask the word from the image. Results are presented in Fig \ref{fig:single_word_completion}. Similar to our earlier findings, the reconstruction quality of single-word completion varies. Some completions are sharp and precise, while others appear blurry. In some few cases, the model produces a sharp reconstruction of an incorrect word (Fig \ref{fig:every}). Unfortunately, due to the blurry nature of many of the results (regardless of their correctness), a quantitative analysis of these results (e.g., by OCRing the reconstructed patch and comparing it to the OCR output of the original patch) is unattainable. 

\begin{figure*}[t]
    \centering
    \begin{subfigure}{0.3\textwidth}
        \centering
        \includegraphics[width=\textwidth, center]{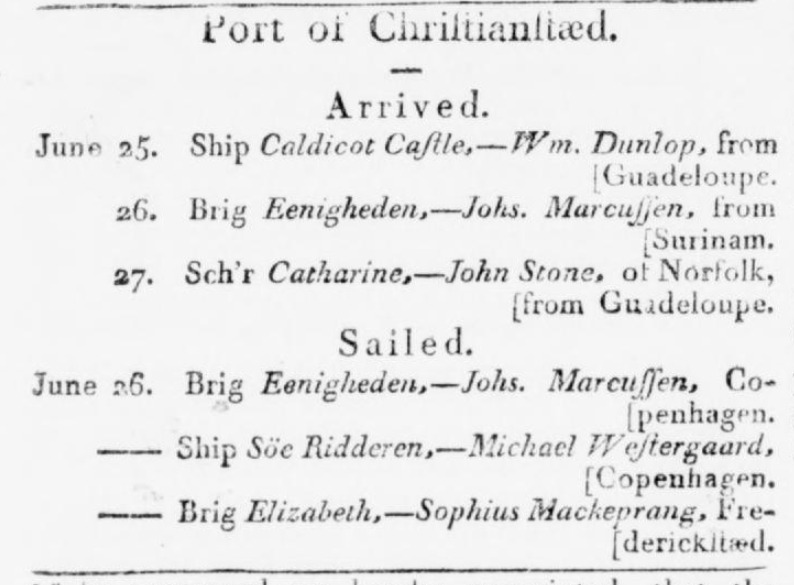}
              \vspace{50pt}
        \caption{Semantic search target.}
        \label{fig:ss_target_2}
      \end{subfigure} 
      \hspace{3pt}
    \begin{subfigure}{0.6\textwidth}
        \centering
        \includegraphics[width=\textwidth, center]{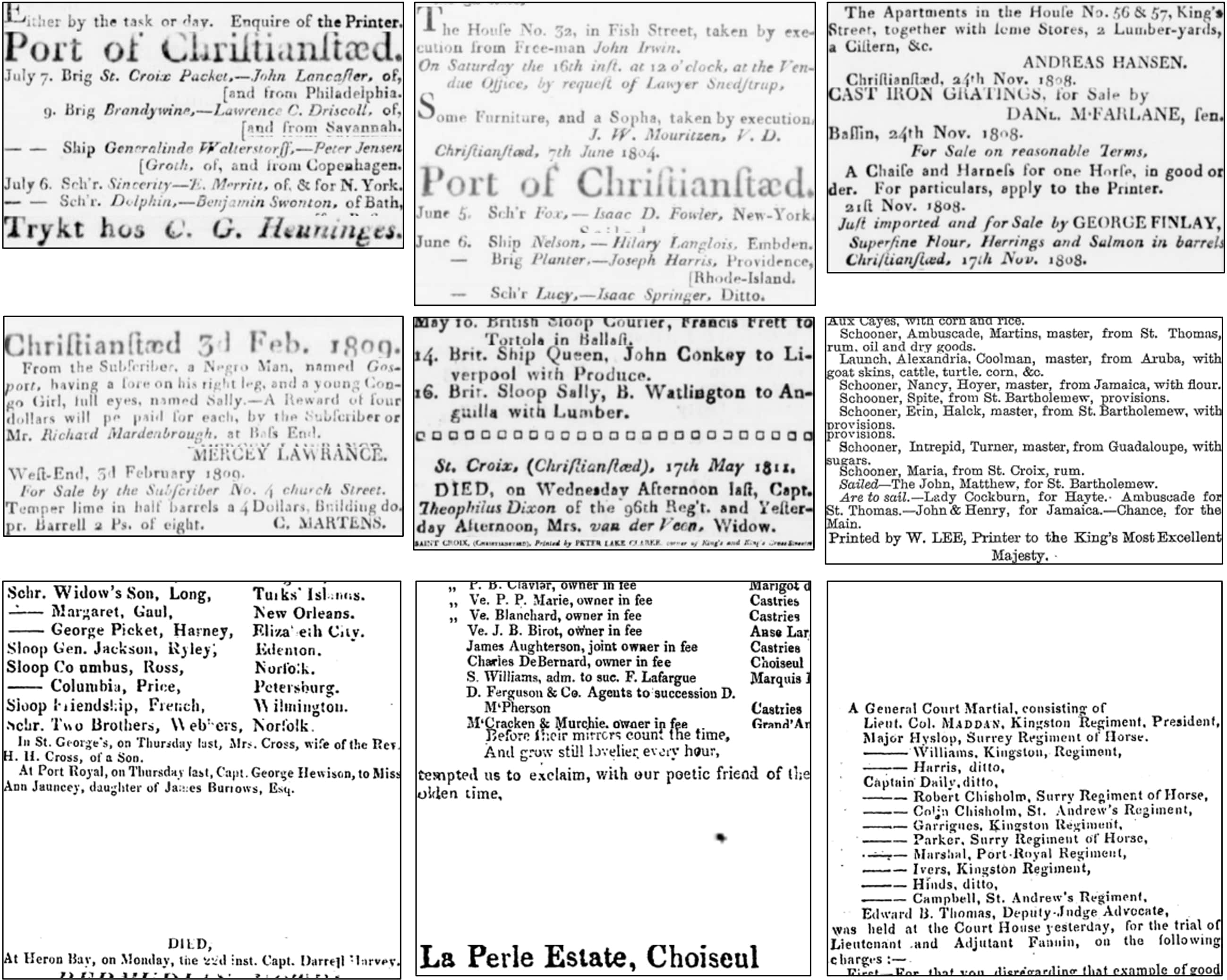}

        \caption{Retrieved scans.}
        \label{fig:ss_res_2}
      \end{subfigure} 

    \caption{Semantic search using our model. (a) is the target of the search, and (b) are scans retrieved from the newspaper corpus.}%
    \label{fig:ss_2}%
\end{figure*}

\textbf{Semantic Search.} A possible useful application of \ourmodel is semantic search. That is, searching in a corpus for historical documents that are semantically similar to a concept of interest. We now analyse \ourmodels ability to assign similar historical scans with similar embeddings. We start by taking a random sample of \num{1000} images from our test set and embed them by averaging the patch embeddings of the final layer of the model. We then reduce the dimensionality of the embeddings with t-SNE \cite{van2008visualizing}. 
Upon visual inspection (Fig \ref{fig:clustering} in App \ref{sec:additional_results}), we see that scans are clustered based on visual similarity and page structure. 

Fig \ref{fig:clustering}, however, does not provide insights regarding the semantic properties of the clusters. Therefore, we also directly use the model in semantic search settings. Specifically, we search our newspapers corpus for scans that are semantically similar to instances of the \textit{Runaways Slaves in Britain} dataset, as well as scans containing shipping ads (See Fig \ref{fig:shipping_ads} in App \ref{sec:additional_results} for examples). To do so, we embed 1M random scans from the corpus. We then calculate the cosine similarity between these embeddings and the embedding of samples from the \textit{Runaways Slaves in Britain} and embeddings of shipping ads. Finally, we manually examine the ten most similar scans to each sample.

Our results (Fig \ref{fig:ss_2} and Fig \ref{fig:ss_1} in App \ref{sec:additional_results}) are encouraging, indicating that the embeddings capture not only structural and visual information, but also the semantic content of the scans. However, the results are still far from perfect, and many retrieved scans are not semantically similar to the search's target. It is highly plausible that additional specialised finetuning (e.g., SentenceBERT's \cite{reimers2019sentence} training scheme) is necessary to produce more semantically meaningful embeddings.

%% file: Modules/5-Finetuning.tex
\everypar{\looseness=-1}

\begin{figure}[t]
    \centering
    \begin{subfigure}{0.42\columnwidth}
        \centering
        \fboxsep=0pt
        \fbox{
        \includegraphics[width=\textwidth, center]{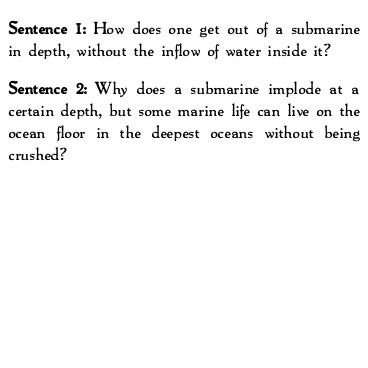}
        }
      \end{subfigure} \hspace{0.5cm}
    \begin{subfigure}{0.42\columnwidth}
        \centering
        \fboxsep=0pt
        \fbox{
        \includegraphics[width=\textwidth, center ]{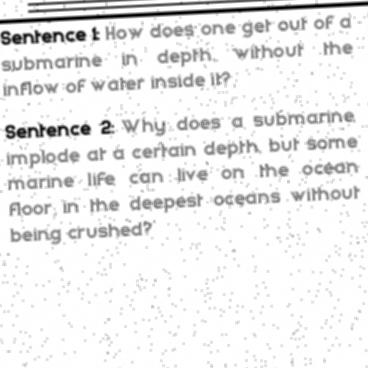}
        }
      \end{subfigure} 

    \caption{Samples from the clean and noisy visual GLUE datasets.}%
    \label{fig:glue_sample}%
\end{figure}

\begin{figure}[t]
    \centering
    \fboxsep=0pt
    \fbox{
        \includegraphics[trim={0.0cm 5.6cm 0.0cm 0.0cm},clip,width=0.75\columnwidth]{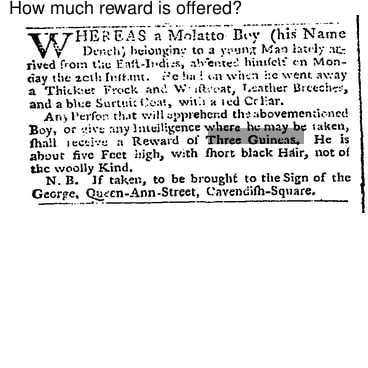}
        }
         \caption{Example from the \textit{Runaways Slaves in Britain} dataset, rendered as visual question answering task. The gray overlay marks the patches containing the answer.}
         \label{fig:runaway_sample}
\end{figure}

\begin{table*}[t]
\centering
\resizebox{\textwidth}{!}{%
\sisetup{table-format = 3.2}
\begin{tabular}{@{}cclSSSSSSSSSS@{}}
\toprule
\multirow{2}{*}{\textbf{Noise}} &
  \multirow{2}{*}{\textbf{Images}} &
  \multirow{2}{*}{\textbf{Model}} &
  \multicolumn{1}{l}{\textbf{MNLI}} &
  \multicolumn{1}{l}{\textbf{QQP}} &
  \multicolumn{1}{l}{\textbf{QNLI}} &
  \multicolumn{1}{l}{\textbf{SST-2}} &
  \multicolumn{1}{l}{\textbf{COLA}} &
  \multicolumn{1}{l}{\textbf{STS-B}} &
  \multicolumn{1}{l}{\textbf{MRPC}} &
  \multicolumn{1}{l}{\textbf{RTE}} &
  \multicolumn{1}{l}{\textbf{WNLI}} &
  \multicolumn{1}{l}{{\multirow{2}{*}{\textbf{AVG}}}} \\
                        &                         &                            & \multicolumn{1}{c}{\num{393}k} & \multicolumn{1}{c}{\num{364}k} & \multicolumn{1}{c}{\num{105}k} & \multicolumn{1}{c}{\num{67}k}  & \multicolumn{1}{c}{\num{8.6}k} & \multicolumn{1}{c}{\num{5.8}k} & \multicolumn{1}{c}{\num{3.7}k} & \multicolumn{1}{c}{\num{2.5}k} & \multicolumn{1}{c}{\num{635}}  &      \\ \midrule
\multirow{5}{*}{\xmark} & \multirow{2}{*}{\xmark} & BERT                       & \textbf{84.1} & \textbf{87.6} & \textbf{91.0} & \textbf{92.6} & \textbf{60.3} & \textbf{88.8} & \textbf{90.2} & \textbf{69.5} & 51.8 & \textbf{80.0} \\
                        &                         & PIXEL                      & 78.5 & 84.5 & 87.8 & 89.6 & 38.4 & 81.1 & 88.2 & 60.5 & 53.8 & 74.1 \\ \cmidrule{2-13}
 & \multirow{3}{*}{\cmark} & CLIP\textsubscript{\emph{lin}} & 50.2 & 64.7 & 67.4 & 79.8 & 4.2  & 56.4 & 74.1 & 51.5 & 25.6 & 52.7 \\
                        &                         & Donut                      & 64.0    & 77.8    & 69.7    & 82.1 & 13.9 & 14.4 & 81.7 & 54.0 & \underline{\textbf{57.7}} & 57.2    \\
                        &                         & \emph{Ours}                       & \underline{70.1} & \underline{82.7} & \underline{82.3} & \underline{82.5} & \underline{15.9} & \underline{80.2} & \underline{83.4} & \underline{59.9} & 54.1 & \underline{67.9} \\ \midrule
\multirow{5}{*}{\cmark}   &   \multirow{5}{*}{\cmark}   & OCR+BERT                   & \textbf{71.7} & 77.5 & \textbf{82.7} & \textbf{85.5} & \textbf{39.7} & 68.4 & \textbf{86.9} & 58.8 & 51.3 & \textbf{69.2} \\
                        &                         & OCR+PIXEL                  & 70.6 & 78.5 & 81.5 & 83.6 & 30.3 & 68.8 & 84.7 & \textbf{59.7} & 58.6 & 68.5 \\ \cmidrule{3-13}
 &  & CLIP\textsubscript{\emph{lin}} & 45.3 & 67.4 & 64.4 & 79.2 & 3.5  & 57.9 & 78.8 & 47.3 & 32.7 & 52.9 \\
                        &                         & Donut                      & 61.6    & 74.1    & 75.1 & 75.5 & 10.2 & 20.6 & 81.9 & 56.7 & \underline{\textbf{60.0}} & 57.3    \\
                        &                         & \emph{Ours}                       & \underline{68.0} & \underline{\textbf{80.4}} & \underline{81.8} & \underline{83.9} & \underline{15.1} & \underline{\textbf{80.4}} & \underline{83.6} & \underline{58.5} & 57.8 & \underline{67.2} \\ \bottomrule
\end{tabular}%
}
\caption{Results for \ourmodel finetuned on GLUE. The metrics are $F_1$ score for QQP and MRPC, Matthew's correlation for COLA, Spearman's $\rho$ for STS-B, and accuracy for the remaining datasets. Bold values indicate the best model in category (noisy/clean), while underscored values indicate the best pixel-based model.}
  \label{tab:glue_results}
\end{table*}

After obtaining a pretrained pixel-based language model adapted to the historical domain (\S\ref{sec:pretraining}), we now move to evaluate its understanding of natural language and its usefulness in addressing historically-oriented NLP tasks. Below, we describe the datasets we use for this and the experimental settings.

\subsection{Language Understanding}
\label{sec:language_understanding}

We adapt the commonly used GLUE benchmark \cite{wang-etal-2018-glue} to gauge our model's understanding of language. We convert GLUE instances into images similar to the process described in \S\ref{sec:fake_scans}. Given a GLUE instance with sentences $s_1, s_2$ ($s_2$ can be empty), we embed $s_1$ and $s_2$ into an HTML template, introducing a line break between the sentences. We then render the HTML files as images.

We generate two versions of this visual GLUE dataset -- clean and noisy. The former is rendered using a single pre-defined font without applying degradations or augmentations, whereas the latter is generated with random fonts and degradations. Fig~\ref{fig:glue_sample} presents a sample of each of the two dataset versions. While the first version allows us to measure \ourmodels understanding of language in ``sterile'' settings, we can use the second version to estimate the robustness of the model to noise common to historical scans. 

\subsection{Historical Question Answering}
\label{sec:historical_qa}

QA applied to historical datasets can be immensely valuable and useful for historians \cite{nadav2023}. Therefore, we assess \ourmodels potential for assisting historians with this important NLP task. We finetune the model on two novel datasets. The first is an adaptation of the classical SQuAD-v2 dataset \cite{squad}, while the second is a genuine historical QA dataset.

\textbf{SQuAD Dataset} We formulate SQuAD-v2 as a patch classification task, as illustrated in Fig \ref{fig:squad_sample} in App \ref{sec:additional_results}. Given a SQuAD instance with question $q$, context $c$ and answer $a$ that is a span in $c$, we render $c$ as an image, $I$ (Fig \ref{fig:visual_historical_squad_just_context}). Then, each patch of $I$ is labelled with $1$ if it contains a part of $a$ or $0$ otherwise. This generates a binary label mask $M$ for $I$, which our model tries to predict (Fig \ref{fig:visual_historical_squad_context_mask_overlay}). If any degradations or augmentations are later applied to $I$, we ensure that $M$ is affected accordingly. Finally, similarly to \citet{lee2022pix2struct}, we concatenate to $I$ a rendering of $q$ and crop the resulting image to the appropriate input size (Fig \ref{fig:visual_historical_squad_final}). 

Generating the binary mask $M$ is not straightforward, as we do not know where $a$ is located inside the generated image $I$. For this purpose, we first use Tesseract to OCR $I$ and generate $\hat{c}$. Next, we use fuzzy string matching to search for $a$ within $\hat{c}$. If a match $\hat{a} \in \hat{c}$ is found, we use Tesseract to find the pixel coordinates of $\hat{a}$ within $I$. We then map the pixel coordinates to patch coordinates and label all the patches containing $\hat{a}$ with $1$. In about 15\% of the cases, Tesseract fails to OCR $I$ properly, and $\hat{a}$ cannot be found in $\hat{c}$, resulting in a higher proportion of SQuAD samples without an answer compared to the text-based version.

As with GLUE, we generate two versions of visual SQuAD, which we use to evaluate \ourmodels performance in both sterile and historical settings.

\textbf{Historical QA Dataset} Finally, we finetune \ourmodel for a real historical QA task. For this, we use the English dataset scraped from the website of the \textit{Runaways Slaves in Britain} project, a searchable database of over $800$ newspaper adverts printed between 1700 and 1780 placed by enslavers who wanted to capture enslaved people who had self-liberated \cite{simon_p_newman_runaway_nodate}. Each ad was manually transcribed and annotated with more than $50$ different attributes, such as the described gender and age, what clothes the enslaved person wore, and their physical description.

Following \citet{nadav2023}, we convert this dataset to match the SQuAD format: given an ad and an annotated attribute, we define the transcribed ad as the context $c$, the attribute as the answer $a$, and manually compose an appropriate question $q$. We process the resulting dataset similarly to how SQuAD is processed, with one key difference: instead of rendering the transcribed ad $c$ as an image, we use the original ad scan. Therefore, we also do not introduce any noise to the images. See \Cref{fig:runaway_sample} for an example instance. We reserve 20\% of the dataset for testing.

\subsection{Training Procedure}
\label{sec:finetuning_procedure}

Similar to BERT, \ourmodel is finetuned for downstream tasks by replacing the decoder with a suitable head. Tab \ref{tab:glue_hyperparameters} in App \ref{app:training} details the hyperparameters used to train \ourmodel on the different GLUE tasks. We use the standard GLUE metrics to evaluate our model. Since GLUE is designed for models of modern English, we use this benchmark to evaluate a checkpoint of our model obtained after training on the artificial modern scans, but before training on the real historical scans. The same checkpoint is also used to evaluate \ourmodel on SQuAD. Conversely, we use the final model checkpoint (after introducing the historical data) to finetune on the historical QA dataset: First, we train the model on the noisy SQuAD and subsequently finetune it on the \textit{Runaways} dataset (see App \ref{app:training} for training details).

To evaluate our model's performance on the QA datasets, we employ various metrics. The primary metrics include binary accuracy, which indicates whether the model agrees with the ground truth regarding the presence of an answer in the context. Additionally, we utilise patch-based accuracy, which measures the ratio of overlapping answer patches between the ground truth mask $M$ and the predicted mask $\hat{M}$, averaged over all the dataset instances for which an answer exists. Finally, we measure the number of times a predicted answer and the ground truth overlap by at least a single patch. We balance the test sets to contain an equal number of examples with and without an answer.  

\subsection{Results}
\label{sec:finetuning_results}

\begin{table}[t]
    \centering
    \resizebox{0.49\textwidth}{!}{%
    \fontsize{10}{10}\selectfont
    \sisetup{table-format = 3.2}
    \begin{tabular}{llcSSS}
    \toprule
    \textbf{Task} &
    \textbf{Model} &
      \multicolumn{1}{c}{\textbf{Noise / Image}} &
      \textbf{\begin{tabular}[c]{@{}c@{}}Binary\\ acc\end{tabular}} &
      \textbf{\begin{tabular}[c]{@{}c@{}}Patch\\ acc\end{tabular}} &
      \multicolumn{1}{c}{\textbf{\begin{tabular}[c]{@{}c@{}}One\\ Overlap\end{tabular}}} \\ \midrule
    \multirow{3}{*}{S} & BERT   & \xmark / \xmark & 72.3 & 47.3 & 53.9 \\ \cmidrule{2-6}
    & \emph{Ours}   & \xmark  / \cmark & 60.3 & 16.4 & 42.2 \\
    & \emph{Ours}     & \cmark / \cmark & 61.7 & 14.4 & 41.2 \\ \midrule
    \multirow{2}{*}{R} & BERT   & - / \xmark & 78.3 & 52.0 & 55.8 \\ \cmidrule{2-6}
    & \emph{Ours} & - / \cmark & 74.7 & 20.0 & 48.8 \\ \bottomrule
    \end{tabular}%
    }
    \caption{Results for \ourmodel finetuned on our visual SQuAD (S) and the \textit{Runaways Slaves} (R) datasets.}
    \label{tab:squad_results}
\end{table}

\textbf{Baselines} We compare \ourmodels performance on GLUE to a variety of strong baselines, covering both OCR-free and OCR-based methods. First, we use CLIP with a ViT-L/14 image encoder in the linear probe setting, which was shown to be effective in a range of settings that require a joint understanding of image and text---including rendered SST-2 \cite{radford-etal-2021-clip}. While we only train a linear model on the extracted CLIP features, compared to full finetuning in \ourmodelnospace, CLIP is about 5$\times$ the size with \circa{427M} parameters and has been trained longer on more data.
Second, we finetune Donut (\S\ref{sec2:pixel_based_models}), which has \circa{200M} parameters and is the closest and strongest OCR-free alternative to \ourmodelnospace. 
Moreover, we finetune BERT and PIXEL on the OCR output of Tesseract. Both BERT and PIXEL are comparable in size and compute budget to \ourmodelnospace. Although BERT has been shown to be overall more effective on standard GLUE than PIXEL, PIXEL is more robust to orthographic noise \cite{rust2022language}. Finally, to obtain an empirical upper limit to our model, we finetune BERT and PIXEL on a standard, not-OCRed version of GLUE. Likewise, for the QA tasks, we compare \ourmodel to BERT trained on a non-OCRed version of the datasets (the \textit{Runaways} dataset was manually transcribed). We describe all baseline setups in App \ref{app:baselines}.

\textbf{GLUE} Tab \ref{tab:glue_results} summarises the performance of \ourmodel on GLUE. Our model demonstrates noteworthy results, achieving scores of above 80 for five out of the nine GLUE tasks. These results serve as evidence of our model's language understanding capabilities. Although our model falls short when compared to text-based BERT by 13 absolute points on average, it achieves competitive results compared to the OCR-then-finetune baselines. Moreover, \ourmodel outperforms other pixel-based models by more than 10 absolute points on average, highlighting the efficacy of our methodology. 



\textbf{Question Answering} According to Tab \ref{tab:squad_results}, our model achieves above guess-level accuracies on these highly challenging tasks, further strengthening the indications that \ourmodel was able to obtain impressive language comprehension skills. Although the binary accuracy on SQuAD is low, hovering around 60\% compared to the 72\% of BERT, the relatively high ``At least one overlap'' score of above 40 indicates that \ourmodel has gained the ability to locate the answer within the scan correctly. Furthermore, \ourmodel displays impressive robustness to noise, with only a marginal decline in performance observed between the clean and noisy versions of the SQuAD dataset, indicating its potential in handling the highly noisy historical domain. The model's performance on the \textit{Runaways Slaves} dataset is particularly noteworthy, reaching a binary accuracy score of nearly 75\% compared to BERT's 78\%, demonstrating the usefulness of the model in application to historically-oriented NLP tasks. We believe that the higher metrics reported for this dataset compared to the standard SQuAD might stem from the fact that \textit{Runaways Slaves in Britain} contains repeated questions (with different contexts), which might render the task more trackable for our model.  

 \begin{figure}[!t]
    \centering
    \begin{subfigure}{0.74\columnwidth}
        \centering
        \includegraphics[trim={0.0cm 5.9cm 0.0cm 0.0cm},clip,width=\textwidth, center]{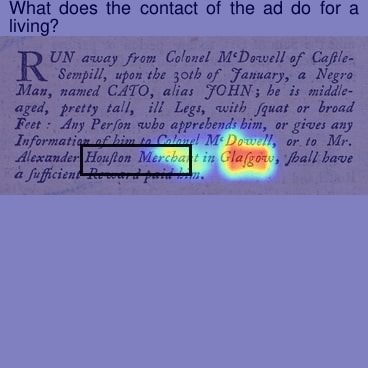}
        \caption{}
        \label{fig:saliency_1}
      \end{subfigure} 
    \begin{subfigure}{0.74\columnwidth}
        \centering
        \includegraphics[trim={0.0cm 6.6cm 0.0cm 0.0cm},clip,width=\textwidth, center]{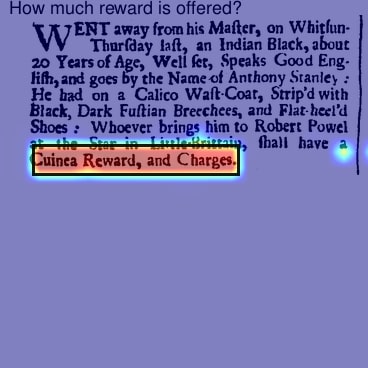}
        \caption{}
        \label{fig:saliency_2}
      \end{subfigure} 

    \caption{Saliency maps of \ourmodel fine-tuned on the \textit{Runaways Slaves in Britain} dataset. Ground truth label in a grey box. The figures were cropped in post-processing.}%
    \label{fig:saliency}%
\end{figure}

\textbf{Saliency Maps} Our patch-based QA approach can also produce visual saliency maps, allowing for a more fine-grained interpretation of model predictions and capabilities \cite{DAS201790}. Fig \ref{fig:saliency} presents two such saliency maps produced by applying the model to test samples from the \textit{Runaways Slaves in Britain} dataset, including a failure case (Fig \ref{fig:saliency_1}) and a successful prediction (Fig \ref{fig:saliency_2}). More examples can be found in Fig~\ref{fig:saliecy_extra} in App \ref{sec:additional_results}.

%% file: Modules/6-Conclusion.tex
In this study, we introduce \ourmodelnospace, an OCR-free language encoder specifically designed for analysing historical documents at the pixel level. We present a novel pretraining method involving a combination of synthetic scans that closely resemble historical documents, as well as real historical newspapers published in the Caribbeans during the 18th and 19th centuries. Through our experiments, we observe that \ourmodel exhibits high proficiency in reconstructing masked image patches, and provide evidence of our model's noteworthy language understanding capabilities. Notably, we successfully apply our model to a historical QA task, achieving a binary accuracy score of nearly 75\%, highlighting its usefulness in this domain. Finally, we note that better evaluation methods are needed to further drive progress in this domain.


%% file: Modules/7-Limitations.tex
We see several limitations regarding our work. First, we focus on the English language only, a high-resource language with strong OCR systems developed for it. By doing so, we neglect low-resource languages for which our model can potentially be more impactful. 

On the same note, we opted to pretrain our model on a single (albeit diverse) historical corpus of newspapers, and its robustness in handling other historical sources is yet to be proven. To address this limitation, we plan to extend our historical corpora in future research endeavours. Expanding the range of the historical training data would not only alleviate this concern but also tackle another limitation; while our model was designed for historical document analysis, most of its pretraining corpora consist of modern texts due to the insufficient availability of large historical datasets.

We also see limitations in the evaluation of \ourmodelnospace. As mentioned in \Cref{sec:pretraining_results}, it is unclear how to empirically quantify the quality of the model's reconstruction of masked image regions, thus necessitating reliance on qualitative evaluation. This qualitative approach may result in a suboptimal model for downstream tasks. Furthermore, the evaluation tasks used to assess our model's language understanding capabilities are limited in their scope. Considering our emphasis on historical language modelling, it is worth noting that the evaluation datasets predominantly cater to models trained on modern language. We rely on a single historical dataset to evaluate our model's performance.

Lastly, due to limited computational resources, we were constrained to training a relatively small-scale model for a limited amount of steps, potentially impeding its ability to develop the capabilities needed to address this challenging task. Insufficient computational capacity also hindered us from conducting comprehensive hyperparameter searches for the downstream tasks, restricting our ability to optimize the model's performance to its full potential. This, perhaps, could enhance our performance metrics and allow \ourmodel to achieve more competitive results on GLUE and higher absolute numbers on SQuAD.

%% file: Modules/8-Appendix.tex
\section{Reproducibility}
\label{app:reproducibility}

\subsection{Training}
\label{app:training}

\begin{table*}[t]
    \centering
    \fontsize{10}{9}\selectfont
    \begin{tabular}{lccccccccc}
        \toprule
        \textbf{Parameter} & \textbf{MNLI} & \textbf{QQP} & \textbf{QNLI} & \textbf{SST-2} & \textbf{COLA} & \textbf{STS-B} & \textbf{MRPC} & \textbf{RTE} & \textbf{WNLI} \\ \midrule
        Classification-head-pooling & \multicolumn{9}{c}{Mean} \\ Optimizer & \multicolumn{9}{c}{AdamW} \\
        Adam $\beta$ & \multicolumn{9}{c}{(\num{0.9}, \num{0.999})} \\
        Adam $\epsilon$ & \multicolumn{9}{c}{\num{1e-8}} \\
        Weight decay & \multicolumn{9}{c}{\num{1e-05}} \\
        Learning rate & \multicolumn{9}{c}{\num{5e-2}}  \\
        Learning rate warmup steps & \multicolumn{9}{c}{\num{100}} \\
        Learning rate schedule & \multicolumn{9}{c}{Cosine annealing} \\
        Batch size & \num{172} & \num{172} & \num{128} & \num{128} & \num{128} & \num{128} & \num{172} & \num{172} & \num{172} \\
        Max steps & \multicolumn{9}{c}{\num{10000}} \\
        Early stopping & \multicolumn{9}{c}{\cmark} \\
        Eval interval (steps/epoch) & \num{500} & \num{500} & \num{500} & \num{500} & \num{100} & \num{100} & \num{100} & \num{250} & \num{100} \\
        Dropout probability & \multicolumn{9}{c}{\num{0.0}} \\ \bottomrule \end{tabular} 
    \caption{The hyperparameters used to train \ourmodel on GLUE tasks.} 
    \label{tab:glue_hyperparameters}
\end{table*}

\paragraph{Pretraining} We pretrain \ourmodel for 1M steps on with the artificial dataset using a batch size of \num{176} (the maximal batch size that fits our system) using AdamW optimizer \cite{kingma2014adam, loshchilov2017decoupled}  with a linear warm-up over the first \num{50}k steps to a peak learning rate of \num{1.5e-4} and a cosine decay to a minimum learning rate of \num{1e-5}. We then train \ourmodel for additional \num{100}k steps with the real historical scans using the same hyperparameters but without warm-up. Pretraining took 10 days on 2 $\times$ 80GB Nvidia A100 GPUs. 

\paragraph{GLUE} Table~\ref{tab:glue_hyperparameters} contains the hyperparameters used to finetune \ourmodel on the GLUE benchmark. We did not run a comprehensive hyperparameter search due to compute limitations; these settings were manually selected based on a small number of preliminary runs.

\paragraph{SQuAD} To finetune \ourmodel on SQuAD, we used a learning rate of \num{6.75e-6}, batch size of \num{128}, dropout probability of \num{0.0} and weight decay of \num{1e-5}. We train the model for \num{50000} steps.

\paragraph{Runaways Slaves in Britain} To finetune \ourmodel on the \textit{Runaways Slaves in Britain} dataset, first trained the model on SQuAD using the hyperparameters mentioned above. Then, we finetuned the resulting model for an additional \num{1000} steps on the \textit{Runaways Slaves in Britain}. The only hyperparameter we changed between the two runs is the dropout probability, which we increased to \num{0.2}.

\subsection{Dataset Generation}
\label{app:dataset_generation}

\paragraph{List of dataset augmentations} To generate the synthetic dataset described in \Cref{sec:fake_scans}, we applied the following transformations to the rendered images: text bleed-through effect; addition of random horizontal and lines; salt and pepper noise; Gaussian blurring; water stains effect; ``holes-in-image" effect; colour jitters on image background; and random rotations.   

\paragraph{Converting the Caribbean Newspapers dataset into 368 $\times$ 368 scans} We convert full newspaper pages into a collection of \num{368} $\times$ \num{368} pixels using the following process.  First, we extract the layout of the page using the Python package Eynollah.\footnote{\url{https://github.com/qurator-spk/eynollah}} This package provides the location of every paragraph on the page, as well as their reading order. As newspapers tend to be multi-columned, we ``linearise'' the page into a single-column document. We crop each paragraph and resize it such that its width equals \num{368} pixels. We then concatenate all the resized paragraphs with respect to their reading order to generate a long, single-column document with a width of \num{368} pixels. Finally, we use a sliding window approach to split the linear page into \num{368} $\times$ \num{368} crops, applying a stride of \num{128} pixels. We reserve 5\% of newspaper issues for validation, using the rest for training. See Fig \ref{fig:real_dataset_samples} in App \ref{sec:additional_results} for dataset examples.

\section{Historical GLUE Baselines}
\label{app:baselines}
For all baselines below, we compute and average scores over 5 random initializations.

\paragraph{OCR + BERT/PIXEL} For each GLUE task, we first generate 5 epochs of noisy training data and run Tesseract on it to obtain noisy text datasets. Similarly, however without oversampling, we obtain noisy versions of our fixed validation sets.
We then finetune BERT-base and PIXEL-base in the same way as \citet{rust2022language}, with one main difference: the noisy OCR output prevents us from separating the first and second sentence in sentence-level tasks. Therefore we treat each sentence pair as a single sequence and leave it for the models to identify sentence boundaries itself, similar to how PHD has to identify sentence boundaries in the images. We use the codebase and training setup from \citet{rust2022language}.\footnote{\url{https://github.com/xplip/pixel}}

\paragraph{CLIP} We run linear probing on CLIP using an adaptation of OpenAI's official codebase.\footnote{\url{https://github.com/openai/CLIP\#linear-probe-evaluation}} We first extract image features from the ViT-L/14 CLIP model and then train a logistic regression model with L-BFGS solver for all classification tasks and an ordinary least squares linear regression model for the regression tasks (only STS-B).

\paragraph{Donut} We finetune Donut-base using an adaptation of ClovaAI's official codebase.\footnote{\url{https://github.com/clovaai/donut}} We frame each of the GLUE tasks as image-to-text tasks: the model receives the (noisy) input image and is trained to produce an output text sequence such as \texttt{<s\_glue><s\_class><positive/> </s\_class></s>}. In this example, taken from SST-2, the {\footnotesize{\texttt{< X >}}} tags are new vocabulary items added to Donut and the label is an added vocabulary item for the positive sentiment class. All classification tasks in GLUE can be represented in this way. For STS-B, where the label is a floating point value denoting the similarity score between two sentences, we follow \citet{raffel-etal-2020-t5} to round and convert the floats into strings.\footnote{Code example in \url{https://github.com/google-research/text-to-text-transfer-transformer/blob/main/t5/data/preprocessors.py\#L816-L855}} We finetune with batch size 32 and learning rate between \num{1e-5} and \num{3e-5} for a maximum of \num{30} epochs or \num{15000} steps on images resized to a resolution of \num{320} $\times$ \num{320} pixels. 

\paragraph{OCR-free BERT/PIXEL} For GLUE, we take results reported in \citep{rust-etal-2021-good}. For SQuAD, we take a BERT model finetuned on SQuAD-v2,\footnote{from \url{https://huggingface.co/deepset/bert-base-cased-squad2}.} and evaluate it on the validation set of SQuAD-v2, after being balanced for the existence of an answer. For the \textit{Runaways Slaves in Britain} dataset, we finetune a BERT-base-cased model\footnote{from \url{https://huggingface.co/bert-base-cased}} on a manually transcribed version of the dataset. We use the default SQuAD-v2 hyperparameters reported in the official Huggingface repository for training on SQuAD-v2.\footnote{\url{https://colab.research.google.com/github/huggingface/notebooks/blob/master/examples/question_answering.ipynb}} We then evaluate the model on a balanced test set, containing 20\% of the ads.

\section{Additional Material}
\label{sec:additional_results}

\begin{figure*}[t]
    \centering
        \includegraphics[width=0.95\textwidth]{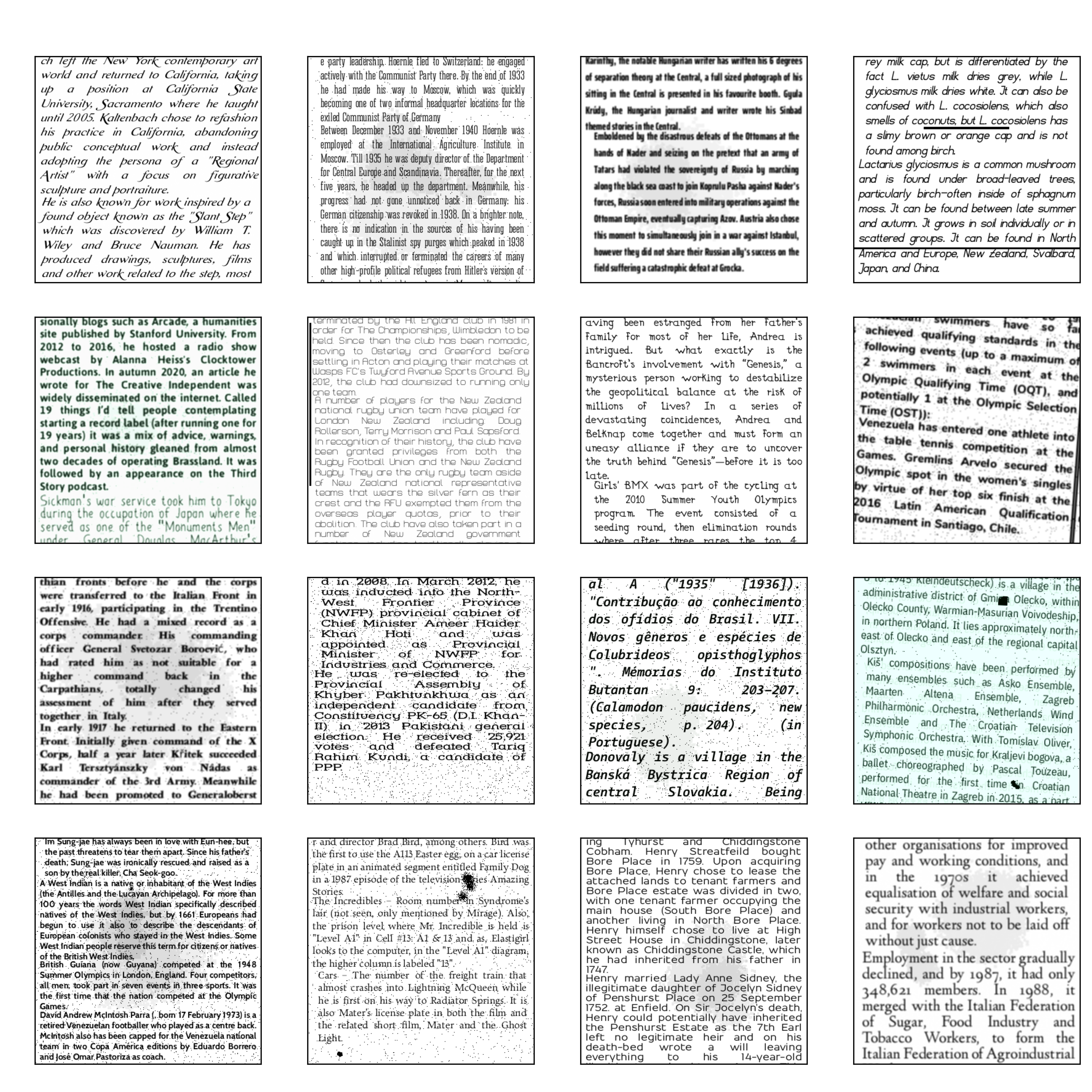}
         \caption{Samples of our artificially generated dataset, and compare to \Cref{fig:real_dataset_samples}.}
         \label{fig:artificial_samples_extra}
\end{figure*}

\begin{figure*}[t]
    \centering
    \begin{subfigure}{0.30\textwidth}
        \centering
        \fboxsep=0pt
         \fbox{
        \includegraphics[width=\textwidth, center]{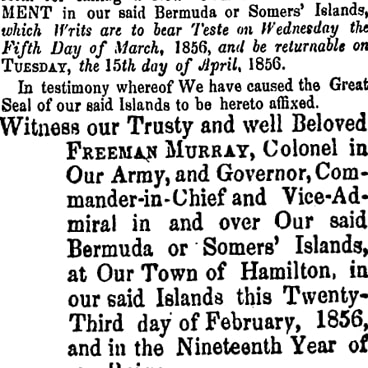}
        }
        \label{fig:caribbean_scans_1}
      \end{subfigure} \hspace{0.1 cm}
    \begin{subfigure}{0.30\textwidth}
        \centering
        \fboxsep=0pt
         \fbox{
        \includegraphics[width=\textwidth, center ]{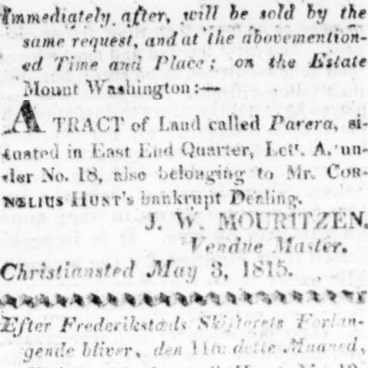}
        }
        \label{fig:caribbean_scans_2}
      \end{subfigure} \hspace{0.1 cm}
        \begin{subfigure}{0.30\textwidth}
        \centering
        \fboxsep=0pt
         \fbox{
        \includegraphics[width=\textwidth, center]{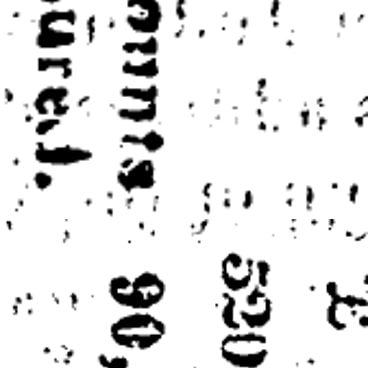}
        }
        \label{fig:caribbean_scans_3}
      \end{subfigure}%
      
\vspace{-0.4 cm}

    \centering
    \begin{subfigure}{0.30\textwidth}
        \centering
        \fboxsep=0pt
         \fbox{
        \includegraphics[width=\textwidth, center]{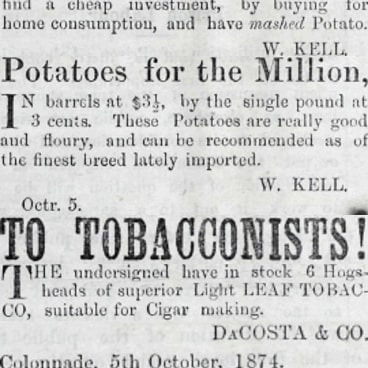}
        }
        \label{fig:caribbean_scans_4}
      \end{subfigure} \hspace{0.1 cm}
    \begin{subfigure}{0.30\textwidth}
        \centering
        \fboxsep=0pt
         \fbox{
        \includegraphics[width=\textwidth, center ]{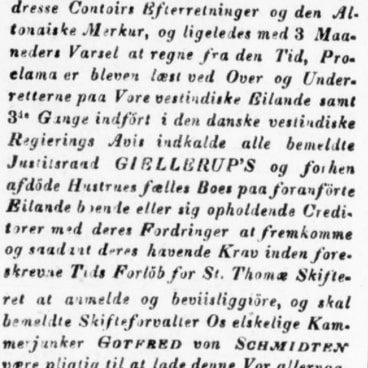}
        }
        \label{fig:caribbean_scans_5}
      \end{subfigure} \hspace{0.1 cm}
        \begin{subfigure}{0.30\textwidth}
        \centering
        \fboxsep=0pt
         \fbox{
        \includegraphics[width=\textwidth, center]{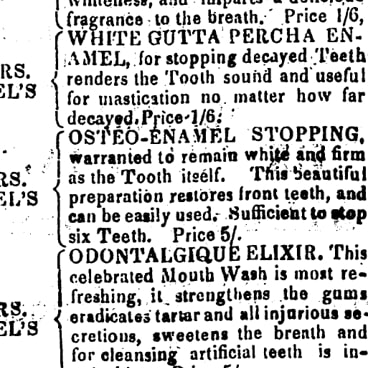}
        }
        \label{fig:caribbean_scans_6}
      \end{subfigure}%
    \caption{Sample scans from the real historical dataset.}%
    \label{fig:real_dataset_samples}%
\end{figure*}

\begin{figure*}[t]
    \centering
    \begin{subfigure}{0.30\textwidth}
        \centering
        \includegraphics[width=\textwidth, center]{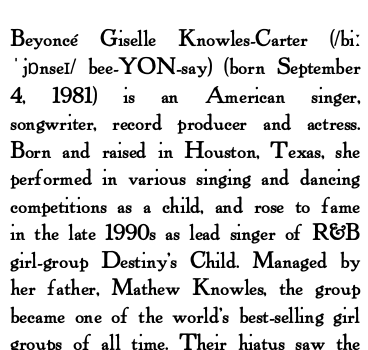}
        \caption{Rendering context $c$ as an image $I$.}
        \label{fig:visual_historical_squad_just_context}
      \end{subfigure} \hspace{0.1 cm}
    \begin{subfigure}{0.30\textwidth}
        \centering
        \includegraphics[width=\textwidth, center ]{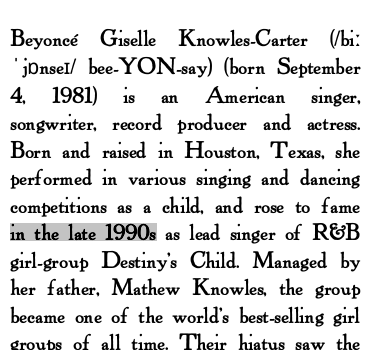}
        \caption{Generating a label mask $M$. }
        \label{fig:visual_historical_squad_context_mask_overlay}
      \end{subfigure} \hspace{0.1 cm}
        \begin{subfigure}{0.30\textwidth}
        \centering
        \includegraphics[width=\textwidth, center, trim={0.0cm 1.3cm 0.0cm 0.0cm},clip]{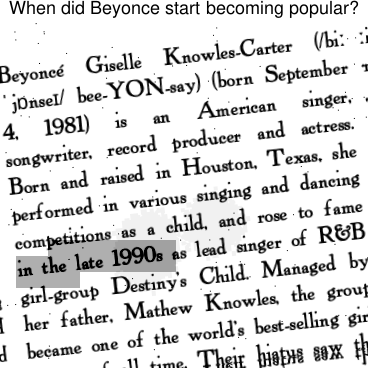}
        \caption{Adding $q$ and degradations.}
        \label{fig:visual_historical_squad_final}
      \end{subfigure}%

    \caption{Process of generating the \textit{Visual SQuAD} dataset. We first render the context as an image (a), generate a patch-level label mask highlighting the answer (b), add noise and concatenate the question (c).}%
    \label{fig:squad_sample}%
\end{figure*}

\begin{figure*}[t]
    \centering
        \includegraphics[width=0.95\textwidth]{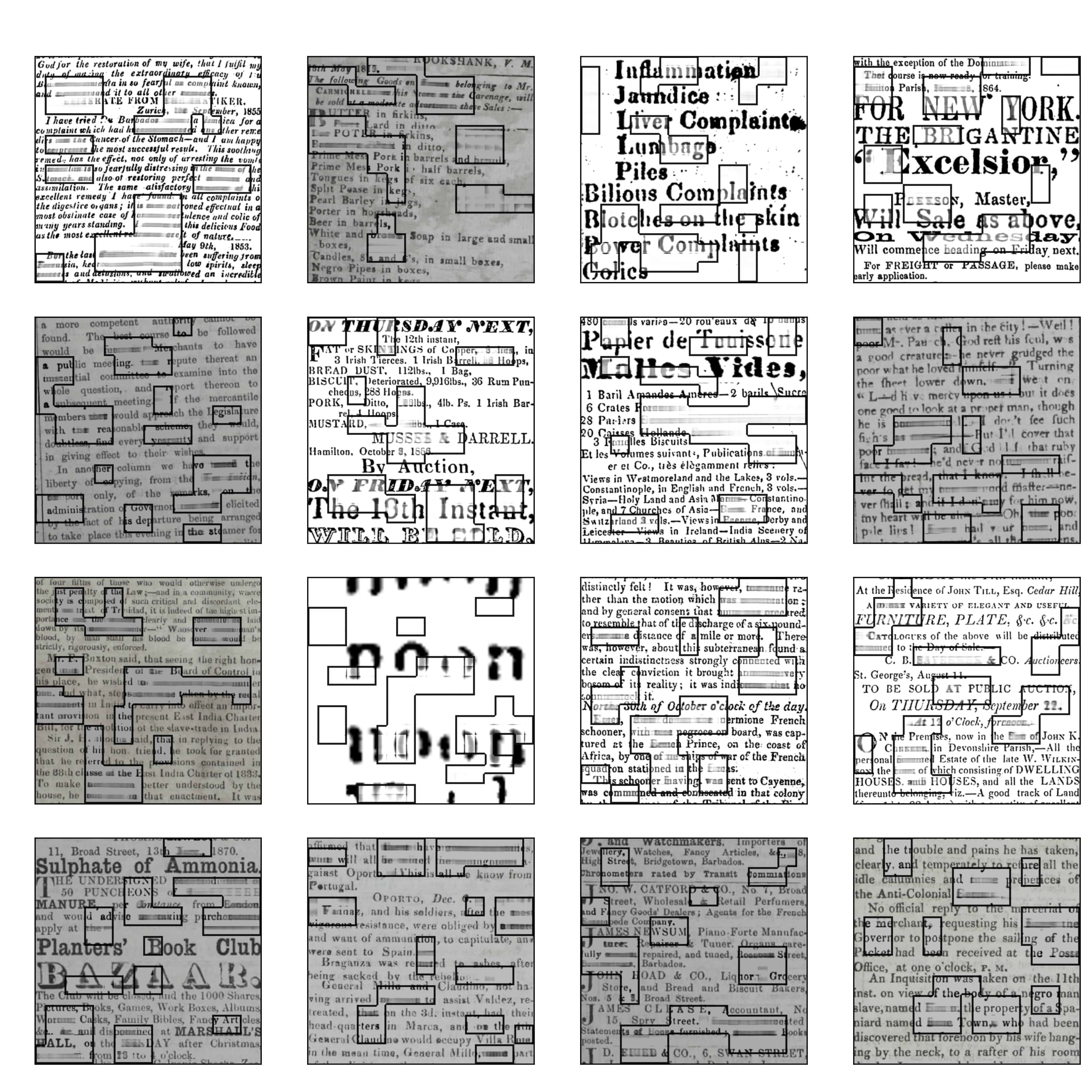}
         \caption{Additional examples of \ourmodels completions.}
         \label{fig:completions_extra}
\end{figure*}

\begin{figure*}[t]
    \centering
    \fboxsep=0pt
    \fbox{
        \includegraphics[width=0.6\textwidth]{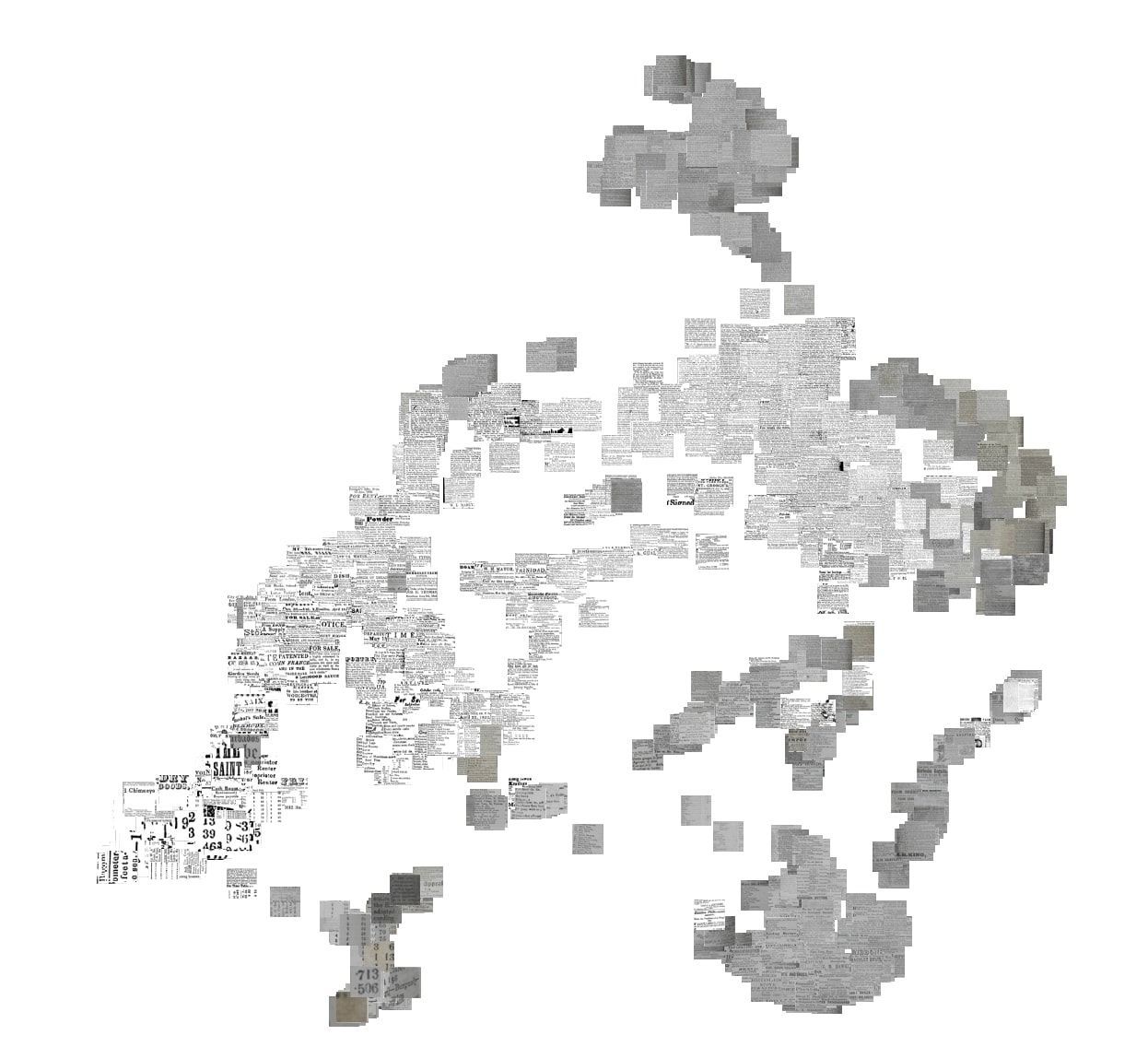}
        }
         \caption{Dimensionality reduction of embedding calculated by our model on historical scans.}
         \label{fig:clustering}
\end{figure*}

\begin{figure*}[t]
    \centering
    \begin{subfigure}{0.3\textwidth}
        \centering
        \includegraphics[width=\textwidth, center]{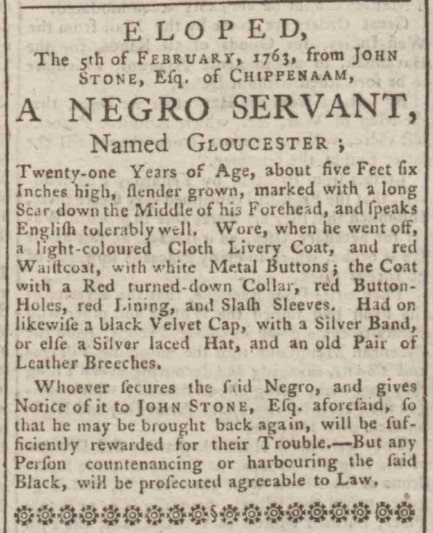}
              \vspace{30pt}
        \caption{Semantic search target.}
        \label{fig:ss_target_1}
      \end{subfigure} 
      \hspace{3pt}
    \begin{subfigure}{0.6\textwidth}
        \centering
        \includegraphics[width=\textwidth, center]{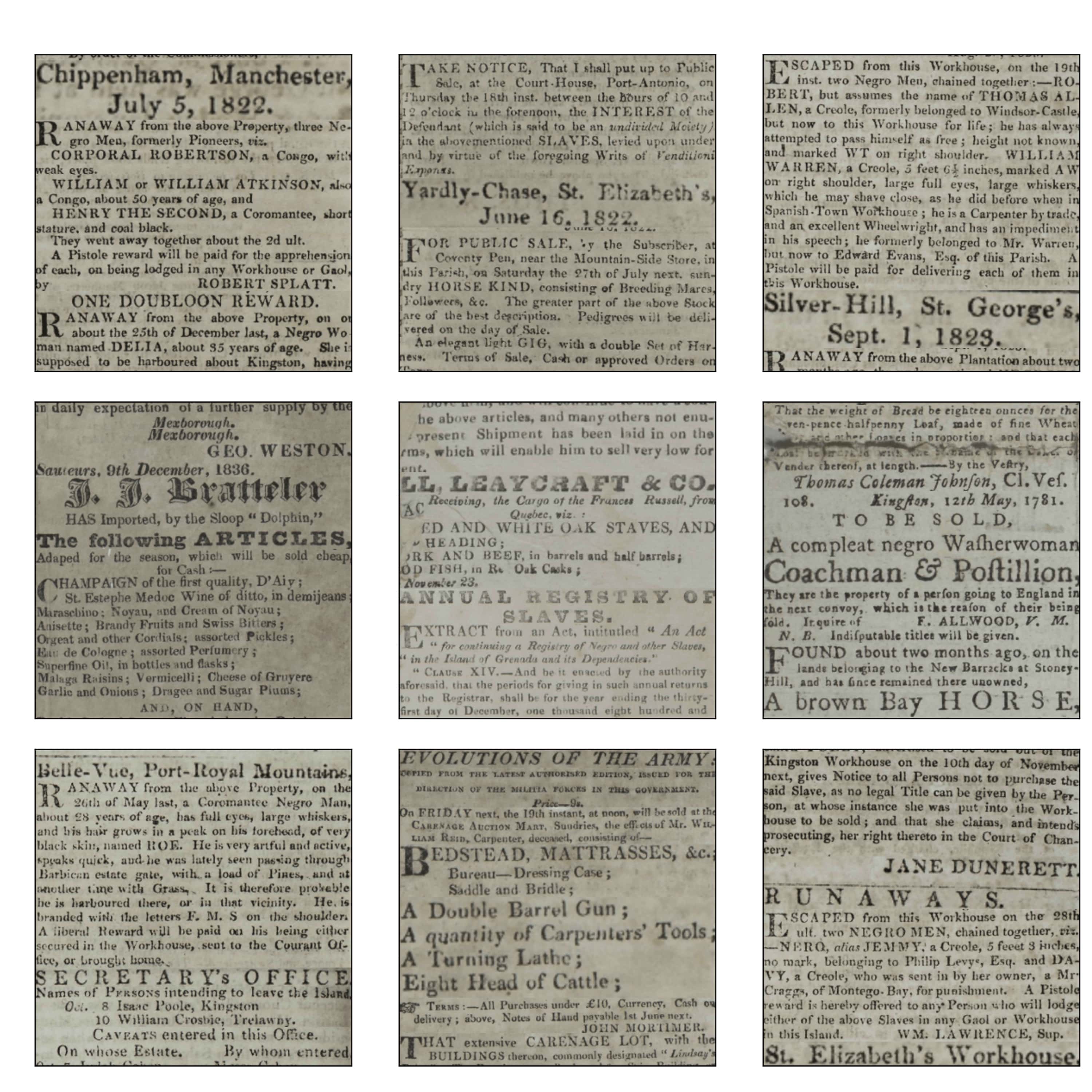}

        \caption{Retrieved scans.}
        \label{fig:ss_res_1}
      \end{subfigure} 

    \caption{Semantic search using our model. (a) is the target of the search, and (b) are scans retrieved from the newspaper corpus.}%
    \label{fig:ss_1}%
\end{figure*}

\begin{figure*}[t]
    \centering
        \includegraphics[width=0.95\textwidth]{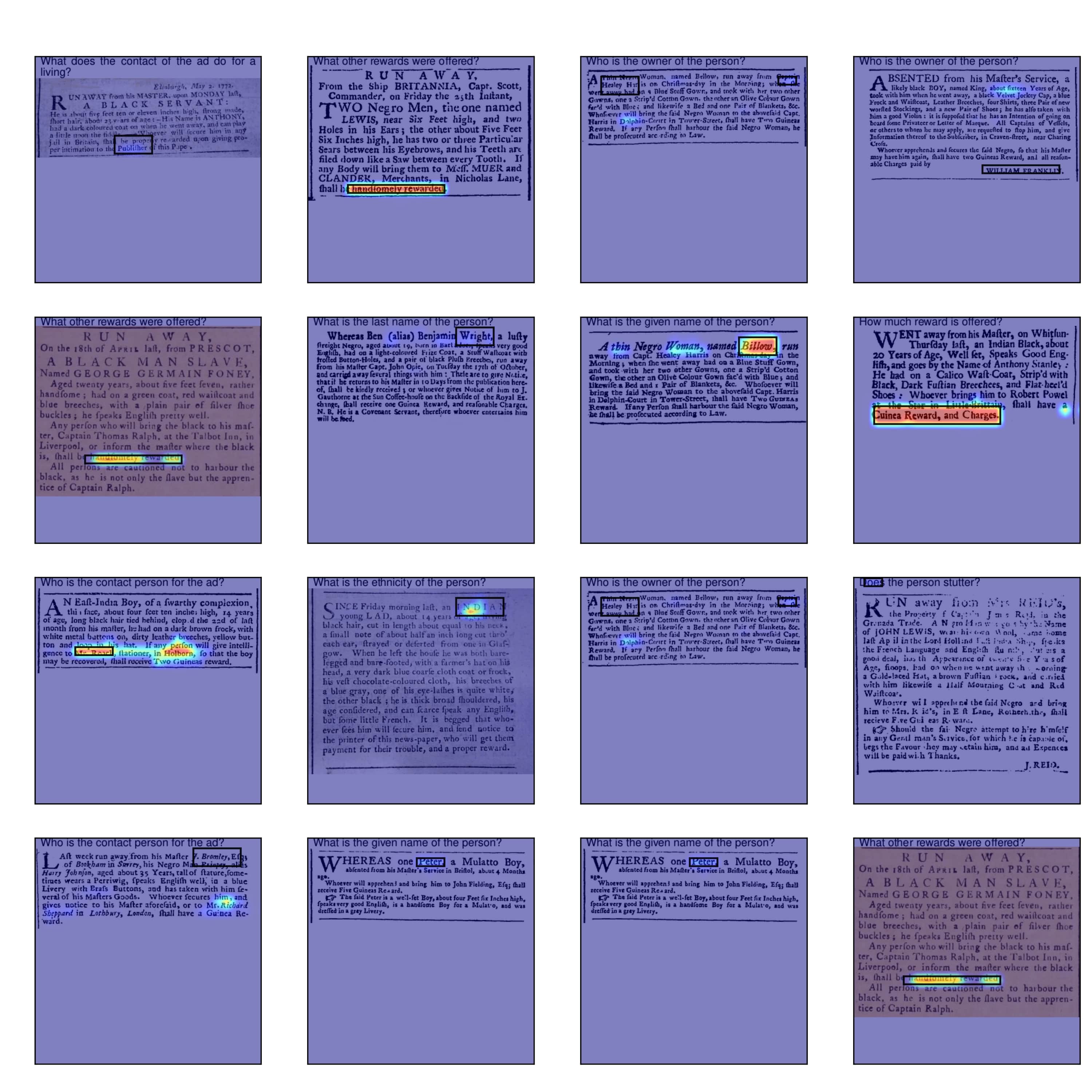}
         \caption{Additional examples of \ourmodels saliency maps for samples from the test set of the \textit{Runaways Slaves in Britain} dataset.}
         \label{fig:saliecy_extra}
\end{figure*}

\begin{figure}
    \centering
    \begin{subfigure}{0.5\columnwidth}
        \centering
        \fboxsep=0pt
        \fbox{
        \includegraphics[width=\textwidth, center]{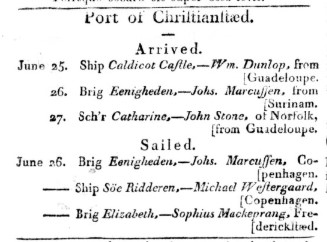}
        }
        
      \end{subfigure} 
      \hspace{3pt}
    \begin{subfigure}{0.38\columnwidth}
        \centering
        \fboxsep=0pt
        \fbox{
        \includegraphics[width=\textwidth, center]{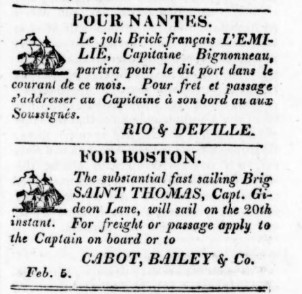}
        }
      \end{subfigure} 

    \caption{Shipping ads samples. Newspapers in the Caribbean region routinely reported on passenger and cargo ships porting and departing the islands. These ads are usually well-structured and contain information such as relevant dates, the ship's captain, route, and cargo.}%
    \label{fig:shipping_ads}%
\end{figure}

\begin{figure}[t]
    \centering
    \begin{subfigure}{0.8\columnwidth}
        \centering
        \fboxsep=0pt
        \fbox{
        \includegraphics[width=\textwidth, center]{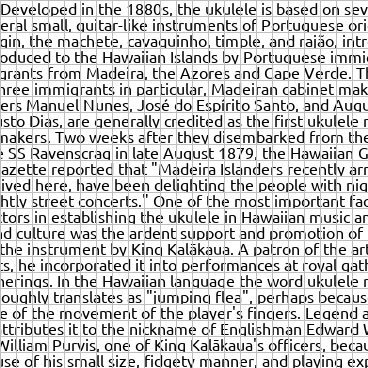}
        }
        \caption{PIXEL's input.}
        \label{fig:pixel_input}
      \end{subfigure} 
      
    \begin{subfigure}{0.8\columnwidth}
        \centering
        \fboxsep=0pt
        \fbox{
        \includegraphics[width=\textwidth, center]{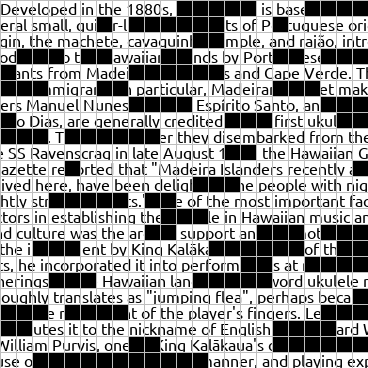}
        }
        \caption{PIXEL's masking.}
        \label{fig:pixel_masking}
      \end{subfigure} 

    \caption{Input samples for PIXEL. The images are rolled, i.e., the actual input resolution is \num{16} $\times$ \num{8464} pixels. The grid represents the \num{16} $\times$ \num{16} patches that the inputs are broken into.}%
    \label{fig:input_samples_with_grid}%
\end{figure}

\begin{figure*}[t]
    \centering
            \fboxsep=0pt
         \fbox{
        \includegraphics[width=0.85\textwidth]{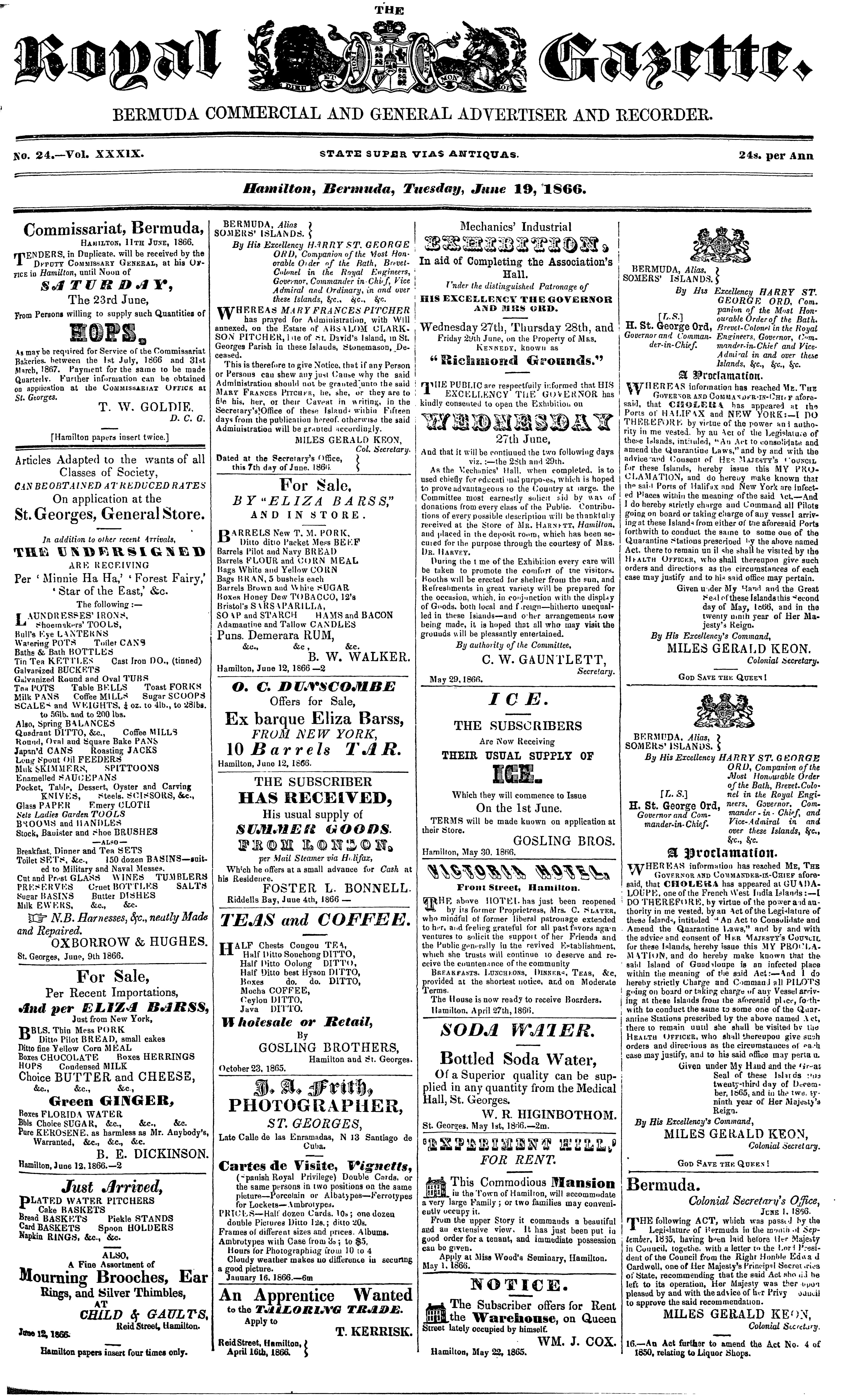}
        }
         \caption{An example of a full newspaper page downloaded from the ``Caribbean project''. \Cref{sec:real_scans} details the way of processing full newspaper pages so that they can be inputted to our model.}
         \label{fig:full_newspaper_page}
\end{figure*}

\noindent\textbf{\Cref{fig:artificial_samples_extra}} additional examples from our artificially generated dataset.

\noindent\textbf{\Cref{fig:real_dataset_samples}} Sample scans from the real historical dataset, as described in \Cref{sec:real_scans}.

\noindent\textbf{\Cref{fig:squad_sample}} The process of generating the \textit{Visual SQuAD} dataset. We first render the context as an image (a), generate a patch-level label mask highlighting the answer (b), add noise and concatenate the question (c).

\noindent\textbf{\Cref{fig:completions_extra}} Additional examples of \ourmodels completions over test set samples.

\noindent\textbf{\Cref{fig:clustering}} Dimensionality reduction of embedding calculated by our model on historical scans. We see that scans are clustered based on visual similarity and page structure. However, further investigation is required to determine whether scans are also clustered based on semantic similarity.

\noindent\textbf{\Cref{fig:ss_1}} Using \ourmodel for semantic search. \Cref{fig:ss_target_1} and is the target of the search (the concept we are looking for), while \Cref{fig:ss_res_1} and are the retrieved scans. 

\noindent\textbf{\Cref{fig:saliecy_extra}} Additional examples of \ourmodels saliency maps for samples from the test set of the \textit{Runaways Slaves in Britain} dataset.

\noindent\textbf{\Cref{fig:shipping_ads}} Examples of shipping ads Newspapers. Newspapers in the Caribbean region routinely reported on passenger and cargo ships porting and departing the islands. These ads are usually well-structured and contain information such as relevant dates, the ship's captain, route, and cargo. 

\noindent\textbf{\Cref{fig:input_samples_with_grid}} Input samples for PIXEL. The images are rolled, i.e., the actual input resolution is 16 $\times$ \num{8464} pixels. The grid represents the 16 $\times$ 16 patches that the inputs are broken into.

\noindent\textbf{\Cref{fig:full_newspaper_page}} An example of a full newspaper page downloaded from the ``Caribbean project''.